\documentclass{svproc}

\usepackage{url}

\usepackage{graphicx}
\usepackage{booktabs}
\usepackage{amsmath}
\usepackage{amsfonts}
\usepackage{hyperref}
\usepackage[a4paper,textwidth=12.2cm,textheight=19.3cm,centering]{geometry}

\begin{document}
\mainmatter            

\title{Accelerating New Product Introduction for Visual Quality Inspection via Few-Shot Diffusion-Based Defect Synthesis}
\titlerunning{Accelerating NPI via Few-Shot Diffusion-Based Defect Synthesis}

\author{Serkan Hamdi G\"u\u{g}\"ul \and Kemal Levi \and Burak Acar}
\authorrunning{S. H. G\"u\u{g}\"ul et al.}

\institute{Relimetrics, Inc.,\\
\email{\{hamdi.gugul, kemal.levi, burak.acar\}@relimetrics.com}}

\maketitle             

\begin{abstract}
Industrial visual inspection systems often suffer from a severe scarcity of labeled defect data, particularly during the early stages of New Product Introduction (NPI). This limitation hinders the deployment of robust supervised detectors precisely when automated quality control is most needed. An end-to-end generative framework is presented for high-fidelity, few-shot defect synthesis that enables both in-domain augmentation and cross-domain transfer. The proposed approach disentangles defect morphology from background appearance by combining masked textual inversion for defect representation learning, noise-blended conditioned generation for surface-aware synthesis, and gradient-aware post-processing for seamless visual integration. The framework is evaluated in two practically relevant settings: few-shot data augmentation, where synthetic samples enrich a small set of real defects, and zero-shot adaptation, where defects learned from a source domain are transferred to a novel target surface without any real target-domain defect examples. Using RF-DETR as the downstream detector, results show that the proposed pipeline reduces the domain gap on a private industrial dataset. In the few-shot setting, synthetic augmentation improves mAP from 78.8\% to 83.3\%. In the zero-shot setting, synthetic domain adaptation improves mAP from 65.0\% to 85.1\%. These results demonstrate that high-fidelity defect synthesis can meaningfully accelerate NPI by enabling effective inspection models before sufficient real defect data has been collected.
\end{abstract}

\section{Introduction}

The performance of modern computer vision systems depends heavily on the availability of large, diverse, and well-annotated datasets \cite{he2016deep}. While such datasets are common in general-purpose recognition problems, industrial quality inspection remains constrained by a persistent rare-event problem: the defects of greatest operational importance are, by definition, infrequent \cite{bergmann2019mvtec}. Hairline scratches, punctures, folds, and related manufacturing anomalies may occur too rarely to support conventional supervised learning pipelines, especially during the early stages of a new production program. As a result, manufacturers often face a critical gap between the need for immediate inspection capability and the delayed availability of representative defect data.

This challenge is particularly acute during New Product Introduction (NPI), where automated inspection systems are expected to operate under severe data scarcity. In such settings, standard augmentation strategies offer only limited benefit because they do not reproduce the structural and textural variability of real defects. What is needed is not merely more data, but realistic, task-relevant defect variation that preserves local morphology while adapting naturally to novel surfaces.

Recent progress in generative modeling has opened a promising path toward this objective. Foundation models and diffusion-based generative methods have shown that synthetic data can enrich downstream learning across various data-scarce environments \cite{azizi2023synthetic,he2023synthetic,shipard2023diversity}. However, industrial inspection imposes stricter requirements than generic image generation. In addition to visual plausibility, useful synthetic defects must exhibit accurate local structure, limited background leakage, and strong compatibility with diverse target materials and lighting conditions.

To address these requirements, an end-to-end, example-driven generative pipeline is presented for few-shot defect synthesis in industrial inspection. The proposed approach leverages latent diffusion models to separate the structural appearance of a defect from its original background, enabling realistic transfer to new surfaces using only three to five annotated reference examples. Specifically, the framework employs masked textual inversion to learn compact defect representations while suppressing background leakage, a noise-blending reverse diffusion process to generate defects conditioned on defect-free target surfaces, and a post-processing stage that aligns color, illumination, and local gradients to improve physical plausibility.
The proposed framework is designed to support two deployment scenarios of high practical importance. In the first, synthetic images are used to augment a limited set of real defect samples in a few-shot regime. In the second, defect concepts learned from an existing source domain are transferred to a new target surface in a zero-shot setting, where no real target-domain defect examples are yet available. These settings directly reflect the operational needs of manufacturers seeking to reduce inspection latency during product launches and process transitions.

The contributions of this work are threefold:
\begin{itemize}
    \item A unified generative pipeline is introduced, tailored to industrial visual inspection, that supports high-fidelity learning and synthesis of localized defects with minimal background contamination.
    \item The resulting synthetic data is demonstrated to improve downstream defect detection in both few-shot and zero-shot settings, yielding substantial gains on a private industrial dataset.
    \item The framework is shown to be lightweight enough to support localized deployment on standard industrial workstations, making it relevant not only as a research concept but also as a practical tool for accelerating NPI.
\end{itemize}

The remainder of the manuscript is organized as follows. Section 2 reviews the related work in generative augmentation and anomaly synthesis. Section 3 reviews the theoretical foundations of diffusion models and personalization methods. Section 4 describes the proposed generative pipeline. Sections 5 and 6 present qualitative and quantitative evaluations, respectively. Section 7 concludes with implications, limitations, and directions for future work.

\section{Related Work}

\subsection{Synthetic Data Augmentation via Diffusion Models}
The utilization of diffusion models for synthetic data augmentation primarily serves as a powerful mechanism for knowledge distillation. Foundational text-to-image models act as massive prior knowledge repositories. By generating synthetic data from these pre-trained models, it is possible to effectively distill broad visual concepts into specialized downstream tasks, treating the generative model as an engine for dataset-driven transfer learning.

Recent literature has validated this paradigm to overcome data scarcity in general computer vision tasks. For example, Azizi et al. \cite{azizi2023synthetic} demonstrated this by fine-tuning large-scale text-to-image diffusion models to produce class-conditional samples. Their evaluations revealed that while training a ResNet-50 \cite{he2016deep} architecture exclusively on synthetic data fell short of real-data baselines (achieving 64.96\% on pure-synthetic data versus 76.81\% accuracy on real data), combining the two domains yielded substantial gains. Specifically, augmenting the real ImageNet training set with synthetic samples increased the ResNet-50 classification accuracy to 79.27\% and pushed the DeiT-B \cite{touvron2021training} model performance from 81.8\% to 84.1\%. 

Building directly on this concept of generative transfer learning, other works have explicitly targeted data-scarce environments. He et al. \cite{he2023synthetic} and Shipard et al. \cite{shipard2023diversity} investigated the viability of using synthetic images from state-of-the-art foundational models to train downstream classifiers when real data is limited. Their combined findings highlight that utilizing the visual diversity of generative data effectively increases the performance of downstream classifiers by providing a broader coverage of the data distribution than what is available in small real datasets.

\subsection{Few-Shot Personalization and Anomaly Generation}
Expanding beyond broad classification, recent methods utilize diffusion models to edit specific semantic features for targeted dataset augmentation. Personalization techniques like Textual Inversion \cite{gal2022image} and DreamBooth \cite{ruiz2023dreambooth} have enabled models to learn novel concepts from as few as three to five images. 

Trabucco et al. \cite{trabucco2023effective} introduced DA-Fusion, which combines Textual Inversion with SDEdit \cite{meng2021sdedit} to perform image-to-image semantic editing. By leveraging pre-trained diffusion models, DA-Fusion can alter high-level class attributes from only a few reference samples. While this effectively increases the semantic variance of natural datasets, its reliance on global semantic editing makes it less suited for the high-precision, localized structural control required to synthesize industrial defects, which are often subtle and highly surface-dependent.

A more specialized approach, and the primary inspiration for the proposed architecture, is Anomaly Diffusion proposed by Hu et al. \cite{hu2023anomaly}. Their mechanism disentangles the anomaly's appearance from its spatial location by combining a learnable embedding with an encoded spatial mask, allowing them to synthesize novel samples on defect-free images for downstream tasks. 

This work adapts this philosophy for cross-domain industrial inspection but introduces several architectural enhancements for streamlined, end-to-end deployment. Unlike previous methods that focus on in-domain augmentation, the proposed approach explicitly addresses the challenge of cross-domain transfer where defects learned from one material are synthesized onto a completely different surface texture. Furthermore, gradient-aware refinement is introduced to eliminate boundary artifacts, which are particularly problematic in industrial inspection where detectors are sensitive to edge discontinuities.

\section{Preliminaries}

\subsection{Latent Diffusion Models (LDMs)}
Early image generation attempts primarily utilized Generative Adversarial Networks (GANs) \cite{goodfellow2014generative} and Variational Autoencoders (VAEs) \cite{kingma2013auto}, which often faced trade-offs between training stability and high-frequency detail. To overcome the inherent trade-offs between training stability in GANs and sample fidelity in VAEs, Denoising Diffusion Probabilistic Models (DDPMs) \cite{ho2020denoising} reframe image generation as an iterative denoising process. However, because DDPMs operate directly in pixel space, they remain computationally prohibitive for high resolution industrial textures. The proposed pipeline utilizes Latent Diffusion Models (LDMs) \cite{rombach2022high}, which perform the diffusion process within the compressed, lower dimensional latent space of a pretrained autoencoder. Given an encoder $\mathcal{E}$ and decoder $\mathcal{D}$, the model operates on latents $z = \mathcal{E}(x)$, minimizing a denoising loss $\mathcal{L}_{LDM}$ conditioned on a text prompt $y$ via a cross-attention mechanism $\tau_\theta(y)$:
\begin{equation}
\mathcal{L}_{LDM} = \mathbb{E}_{z, y, \epsilon, t} [ \| \epsilon - \epsilon_\theta(z_t, t, \tau_\theta(y)) \|^2 ]
\end{equation}
By shifting the computation to the latent space, a high-fidelity synthesis engine is achieved that remains compatible with localized industrial workstations without requiring server-grade infrastructure.

\subsection{Personalization Methods}
To steer the extensive prior knowledge of large-scale pre-trained models, personalization techniques learn novel concepts from a small reference set. Strategies like DreamBooth \cite{ruiz2023dreambooth} fine-tune the entire network weights, while Low-Rank Adaptation (LoRA) \cite{hu2021lora} injects trainable low-rank decomposition matrices into the architecture. In this work, Textual Inversion \cite{gal2022image} is adopted for its extreme modularity and structural fidelity. By optimizing a new word embedding $v^*$ in the text encoder's continuous space while keeping the generative model frozen, reference images are reconstructed by minimizing:
\begin{equation}
v^* = \arg\min_v \mathbb{E}_{z, \epsilon, t} [ \| \epsilon - \epsilon_\theta(z_t, t, \tau_\theta(S_*(v))) \|^2 ]
\end{equation}
This allows the proposed framework to represent unique defects as compact files (only takes a few KBs). This makes it possible to store and deploy many different defect types instantly without the storage overhead of duplicating large model weights. Crucially, synthetic images generated via this method preserve the subtle morphological features of the original real samples, making them effective for training downstream models for tasks such as defect detection.

\subsection{Localized Defect Synthesis via Masked Inpainting}
Modifying only specific regions while preserving a global defect-free background is critical for industrial realism. A solution inspired by RePaint \cite{lugmayr2022repaint} and Blended Diffusion \cite{avrahami2022blended} is utilized. The unmasked regions $m$ are anchored to the original background using a forward-diffused (noised) sample $q(x_{t-1} | x_0)$, while allowing the model's reverse prediction $p_\theta$ to freely generate the defect within the mask:
\begin{equation}
x_{t-1} = m \odot q(x_{t-1} | x_0) + (1 - m) \odot p_\theta(x_{t-1} | x_t)
\end{equation}
This ensures that the generated defect is structurally and contextually adapted to the new surface's texture and lighting.

\section{Methodology}

\begin{figure}[t]
    \centering
    \includegraphics[width=1\textwidth]{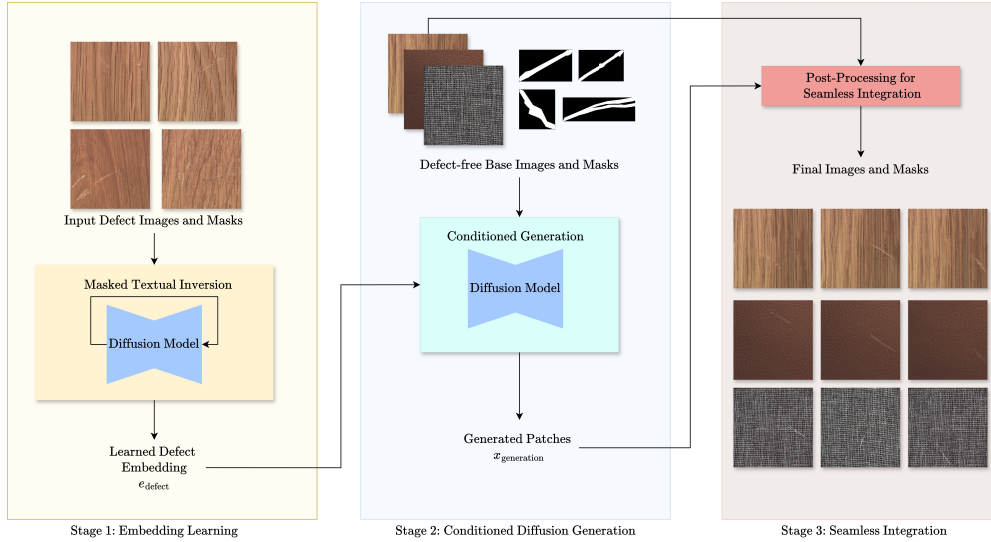}
    \caption{The three-stage generative pipeline: Embedding Learning, Conditioned Generation, and Seamless Integration.}
    \label{fig:schema}
\end{figure}

The proposed generative pipeline for industrial defect synthesis is structured into three primary stages: (i) defect embedding learning via masked textual inversion, (ii) conditioned diffusion generation with dynamic Classifier-Free Guidance (CFG) scheduling, and (iii) sophisticated statistical and gradient-aware blending. This modular architecture allows the system to learn a defect concept from a source domain and accurately synthesize it onto heterogeneous target domains, as illustrated in Figure~\ref{fig:schema}.

The first stage of the pipeline captures the unique visual characteristics of a specific defect from a minimal dataset where 3 to 5 samples are enough. To ensure high-fidelity learning and prevent the loss of small defects during resizing, a centering and patch-extraction pre-processing step is implemented. For each sample, the bounding box of the defect mask is calculated, and a centered patch is extracted with the resolution of the LDM encoder, ensuring the defect remains the focal point with maximum pixel density. While standard Textual Inversion \cite{gal2022image} optimizes a pseudo-word embedding $v^*$ using a global reconstruction loss, a Masked Diffusion Loss is used to disentangle the defect from its original background. Let $\epsilon_\theta(z_t, t, v)$ be the pre-trained LDM. For a latent representation $z_t$ and its corresponding downsampled binary mask $M$, the objective is:
\begin{equation}
v^* = \arg\min_v \mathbb{E}_{t,\epsilon,y} [ \| M \odot (\epsilon - \epsilon_\theta(z_t, t, v)) \|^2 ]
\end{equation}
where $\odot$ denotes the Hadamard product. This restriction ensures that the gradient with respect to the embedding is derived purely from the defect pixels, preventing background texture leakage. Omitting this masked constraint leads to severe background texture leakage into the optimized embedding. The embedding is trained using the AdamW optimizer with an Exponential Moving Average (EMA) decay of 0.995.

In the second stage, the learned embedding $v^*$ is used to synthesize the defect onto a target defect-free background $y_{base}$ within a specified mask $M_{desired}$. A noise-blending reverse diffusion process inspired by RePaint \cite{lugmayr2022repaint} is utilized. At each reverse timestep $t \in \{T, \dots, 1\}$, the latent $z_{t-1}$ is constructed by blending the generative path (conditioned on $v^*$) with a preservation path:
\begin{equation}
z_{t-1} = M \odot z_{t-1}^{gen} + (1 - M) \odot q(z_{t-1} | z_0^{base})
\end{equation}
This ensures that the area outside the mask remains anchored to the base image while the defect evolves within the mask. To optimize visual fidelity, a sigmoid-annealed CFG schedule is introduced. Unlike traditional constant guidance, the guidance scale transitions from a high guidance value at early steps to a low guidance value at late steps. Omitting this dynamic scheduling in favor of a constant high CFG results in rigid, exaggerated defect structures that fail to adapt to the target surface texture. This trajectory establishes a globally coherent structural anchor before softening the edges to naturally harmonize with the base image's texture and grain.

To eliminate remaining visual discontinuities, the final stage applies post-processing in the CIE LAB color space. Unlike RGB, LAB decouples Luminance ($L$) from chromatic components ($a, b$), allowing lighting to be harmonized without distorting color. Scaled CDF Matching is employed to align the global statistical properties of the generated patch with the target surface. Finally, Poisson blending \cite{perez2003poisson} is applied to match the gradients of the synthesized patch while enforcing boundary conditions, ensuring the synthetic defect appears physically integrated into the target texture. Ablating this post-processing stage results in noticeable color and lighting shifts inherited from the latent-space noise of the diffusion model, alongside edge artifacts and boundary seams.

In addition to its generative performance, the proposed pipeline operates within modest hardware requirements, making it accessible for localized factory deployment. The initial defect concept learning phase requires approximately 16 hours on a consumer-grade NVIDIA RTX 5080 GPU, operating with a peak memory footprint of under 8 GB VRAM. For standard edge server deployments, this learning phase requires approximately 24 hours on a server-grade NVIDIA L4 GPU. Inference is fast, with a single high-fidelity synthetic defect generated in approximately 15 seconds. Furthermore, the downstream detector (RF-DETR) is trained in under 1 hour on the RTX 5080 workstation, and achieves a real-time inference throughput of over 60 FPS. This low resource overhead ensures that the system can be effectively scaled and deployed on standard industrial workstations without requiring expensive, server-grade computing infrastructure.

\begin{figure}[t]
    \centering
    \includegraphics[width=\textwidth]{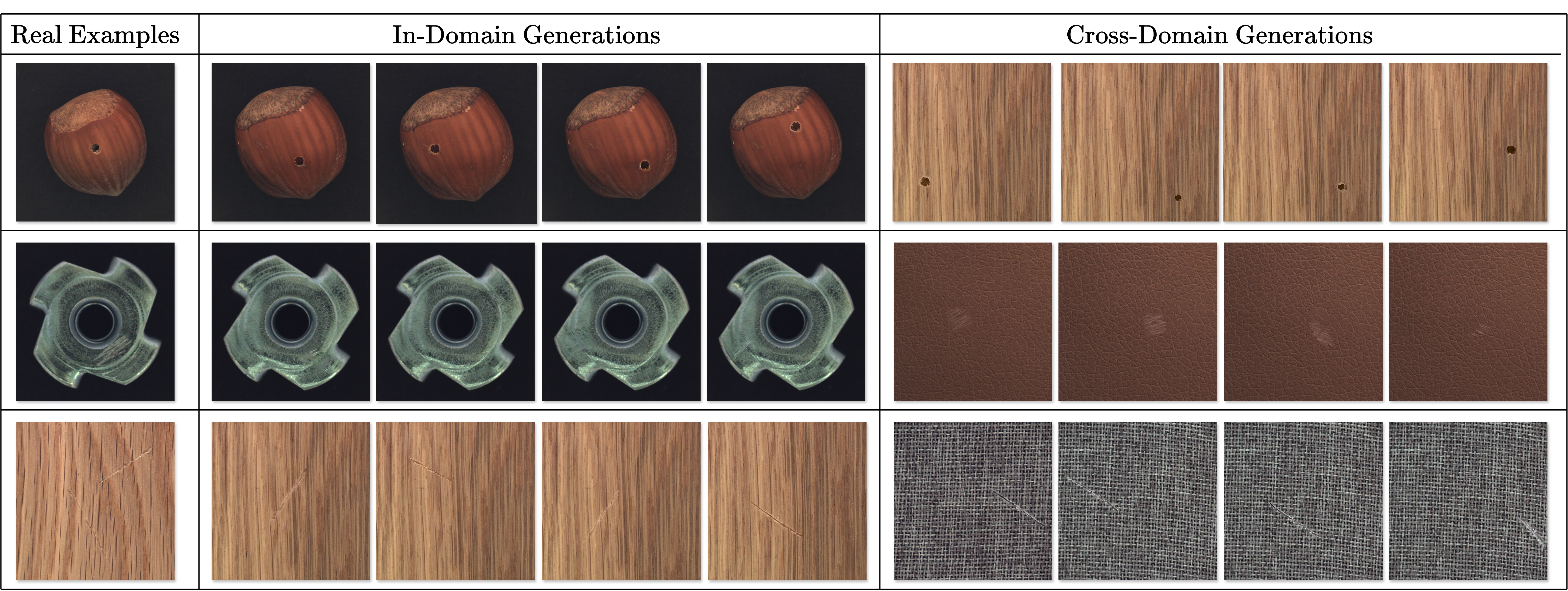}
    \caption{Qualitative defect synthesis and domain transfer results (MVTec AD \cite{bergmann2019mvtec}). \textbf{Left:} Real defect. \textbf{Middle (In-Domain):} Synthesized defects on native source material. \textbf{Right (Cross-Domain):} Zero-shot transfer onto unseen material textures, demonstrating successful disentanglement of defect morphology from background context.}
    \label{fig:domain_generations}
\end{figure}

\section{Qualitative Analysis: Visual Fidelity and Domain Transfer}

To evaluate the generative capabilities of the pipeline, the visual fidelity of the synthesized defects is analyzed using standard industrial inspection textures from the MVTec AD dataset \cite{bergmann2019mvtec}. The primary objective is to verify that the generated defects are not only structurally coherent but also photorealistically consistent with their surrounding backgrounds, which is critical for training robust downstream detection models.

In-domain generations are first assessed, where the defect is synthesized onto defect-free instances of the same material class used for embedding learning. For example, a scratch concept learned from a metal nut is generated onto novel, clean metal nuts. As shown in Figure~\ref{fig:domain_generations} (row 2), the pipeline preserves the high-frequency morphological details while naturally conforming to the local texture of the target background.

The core operational utility of the system lies in \textit{cross-domain} generation, which facilitates zero-shot adaptation for new production lines. In this setup, a defect embedding learned from one material (e.g., wood) is applied to a completely different texture (e.g., metal or leather). The results in Figure~\ref{fig:domain_generations} (Right) demonstrate that the synthesized defects adopt the visual characteristics of the new target domain, such as the specularity of metal or the grain of leather while maintaining the underlying geometry of the original defect. 

The synthesized defects seamlessly adopt the visual characteristics of the new target domains, resulting in realistic generations free of traditional boundary artifacts. This confirms that the defect's morphology has been successfully disentangled from its original context.

\section{Experimental Results: Accelerating New Product Introduction}

To evaluate the impact of the proposed generative pipeline on the manufacturing lifecycle, two distinct experiments are conducted simulating the challenges of new product introduction (NPI). The first experiment assesses few-shot synthetic data augmentation, where a small number of real defect samples are available for the target domain. The second experiment evaluates zero-shot detection, mirroring a scenario where no real defect data exists for a new production line. 

In both experiments, the primary evaluation metrics are mean average precision (mAP) and the best F1-score. Operating precision and recall are also reported at a 40\% confidence threshold, reflecting the default operating point of the inspection software. The defects generated and detected in the experiments are scratches, and since industrial scratches are exceptionally thin and lack axis alignment, even minor bounding box deviations can cause severe drops in traditional Intersection over Union (IoU) scores. Therefore, mAP@0.01 is emphasized, which offers a more reliable assessment of the model's ability to detect these challenging defects. 

\subsection{Few-Shot Synthetic Data Augmentation}

In this experiment, a private industrial dataset is utilized, as it provides real-world production-line defects with bounding box annotations that are not available in public anomaly detection benchmarks primarily designed for unsupervised segmentation tasks. While the generative pipeline is designed to operate with a minimal set of 3--5 reference images, exactly 4 real scratch samples are selected for these evaluations as a representative value within this range. The evaluation examines whether generative AI-based synthetic data augmentation can improve detection performance by introducing additional structural variations. 
The generative pipeline is leveraged to learn an embedding from these 4 samples and to synthesize 150 unique high-fidelity defect images on 20 real defect-free backgrounds of the new product (also called Product B in the second experiment). For evaluation, a separate test set of 129 real scratch images from the new product is used, which does not include the 4 images used for embedding learning.

 \begin{table}[h]
    \centering
    \caption{Overview of the Few-Shot model configurations (Experiment 1).}
    \label{tab:few_shot_configurations}
    \renewcommand{\arraystretch}{1}
    \begin{tabular}{l l l c c}
        \hline
        Model Name & Training Dataset & Base Weights & Epochs & LR \\
        \hline
        Real Only & Real Data & Pretrained & 40 & 1e-4 \\
        Synthetic Only & Synthetic Data & Pretrained & 40 & 1e-4 \\
        Real+Synthetic & Real + Synthetic Data & Pretrained & 40 & 1e-4 \\
        \hline
    \end{tabular}
\end{table}

\begin{figure}[h]
    \centering
    \includegraphics[width=0.8\linewidth]{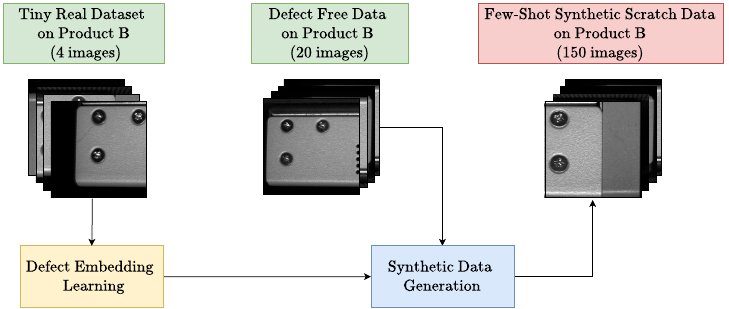}
    \caption{Flow for Few-Shot Synthetic Data Generation: Learning a defect embedding from 4 real Product B samples to generate 150 diverse synthetic variations on the same product surface.}
    \label{fig:few_shot_experiment_pipeline}
\end{figure}

To evaluate the impact of augmentation, three model configurations are investigated. The Real Only model is trained using the four real scratch samples, which are duplicated to reach a total dataset size of 300 images. This duplication ensures that all models maintain the same computational budget per epoch and allows for the application of standard RF-DETR \cite{robinson2025rfdetr} random augmentations to the duplicated samples. The Synthetic Only model is trained exclusively on the 150 synthetic images generated by the proposed pipeline, similarly up-sampled to 300 images to maintain training parity. Finally, the Real+Synthetic model is trained on a combination of 150 real duplicates and 150 synthetic images, resulting in a 300-image training set. A target size of 300 was chosen to integrate exactly 150 synthetic images while maintaining a 1:1 real-to-synthetic ratio; internal testing indicates that this ratio provides the best detection results for joint training. The few-shot synthetic data generation process is illustrated in Figure~\ref{fig:few_shot_experiment_pipeline} and models to be evaluated are detailed in Table~\ref{tab:few_shot_configurations}.

\begin{figure}[t]
    \centering
    \includegraphics[width=1\textwidth]{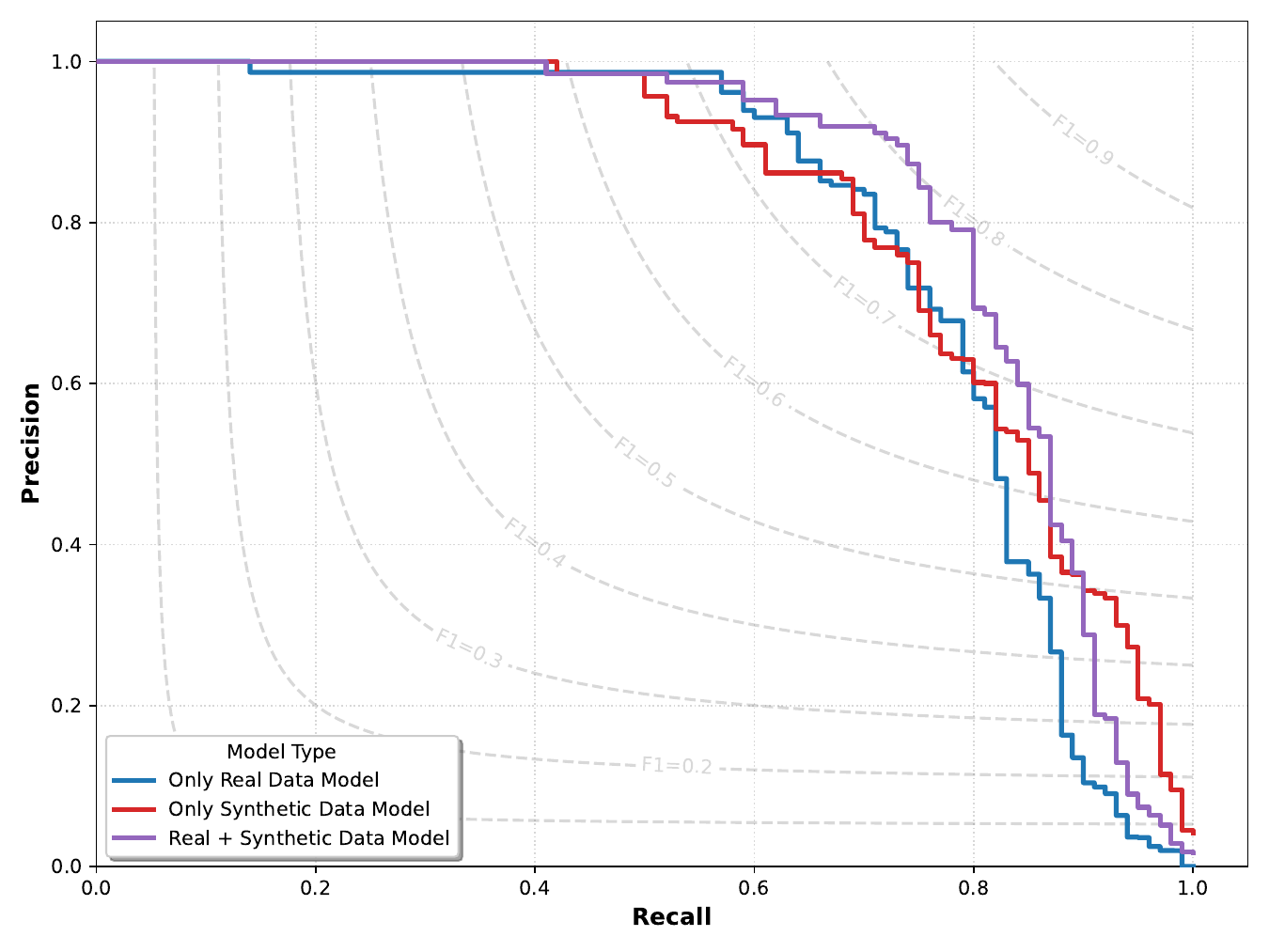}
    \caption{Precision-Recall curves for Few-Shot synthetic data augmentation experiment models.}
    \label{fig:pr_curve_few}
\end{figure}

\begin{table}[t]
    \centering
    \caption{Few-Shot Synthetic Data Augmentation Experiment Performances at IoU@0.01 and Operating Point (Confidence Threshold of 40\%)}
    \label{tab:few_shot_metrics}
    \renewcommand{\arraystretch}{1.2}
    \begin{tabular}{l c c c c c}
        \hline
        & \multicolumn{2}{c}{Overall Performance} & \multicolumn{3}{c}{Operating Point (40\%)} \\
        \cmidrule(lr){2-3} \cmidrule(lr){4-6}
        Model Name & mAP@0.01 & Best F1 & Precision & Recall & F-1 \\
        \hline
        Real Only & 78.8\% & 76.5\% & \textbf{98.6\%} & 50.4\% & 66.7\% \\
        Synthetic Only & 81.2\% & 76.1\% & 98.5\% & 45.7\% & 62.4\% \\
        Real+Synthetic & \textbf{83.3\%} & \textbf{80.9\%} & 93.3\% & \textbf{64.3\%} & \textbf{76.1\%} \\
        \hline
    \end{tabular}
\end{table}

The results in Table~\ref{tab:few_shot_metrics} and Figure~\ref{fig:pr_curve_few} illustrate the advantages of generative augmentation in a few-shot industrial setting. Initially, it is observed that the Real Only baseline achieves a respectable 78.8\% mAP, demonstrating that the SOTA RF-DETR architecture is capable of establishing a functional baseline even with a minimal set of four images. However, the training data can be enriched for a better detector using synthetic augmentation with the proposed generative pipeline, effectively steering the general knowledge of the diffusion model toward this specific industrial use case. 

Specifically, the Real+Synthetic model achieves 83.3\% mAP (as shown by the Precision-Recall curves in Figure~\ref{fig:pr_curve_few}), representing a 4.5-point improvement over the baseline and an increase in operating Recall (64.3\% vs. 50.4\%) at the 0.4 confidence threshold. The Synthetic Only results (81.2\% mAP) are included primarily to showcase the reliability and high fidelity of the synthetic data, which proves capable of training a high-performing model even in total isolation. In a production environment, the combined Real+Synthetic strategy represents the preferred deployment mode, as it effectively anchors the synthetic diversity to the available ground-truth data.

\subsection{Zero-Shot Detection with Synthetic Data}

The second experiment simulates a pure zero-shot scenario where no real defect samples are available for the target surface, denoted as Product B. To accelerate deployment, samples of the same defect type from an existing production line, also called Product A, are utilized to bridge the domain gap. 

\begin{table}[h]
    \centering
    \caption{Overview of the Zero-Shot model configurations (Experiment 2).}
    \label{tab:zero_shot_configurations}
    \renewcommand{\arraystretch}{1}
    \begin{tabular}{l l l}
        \hline
        Model Name & Training Dataset & Base Weights \\
        \hline
        Product A Model & Real Data (Product A) & Pretrained \\
        Product B Synthetic Model & Synthetic Data (Product B) & Pretrained \\
        Product B Synthetic Finetuning Model & Synthetic Data (Product B) & Product A Model \\
        \hline
    \end{tabular}
\end{table}

\begin{figure}[t]
    \centering
    \includegraphics[width=0.8\linewidth]{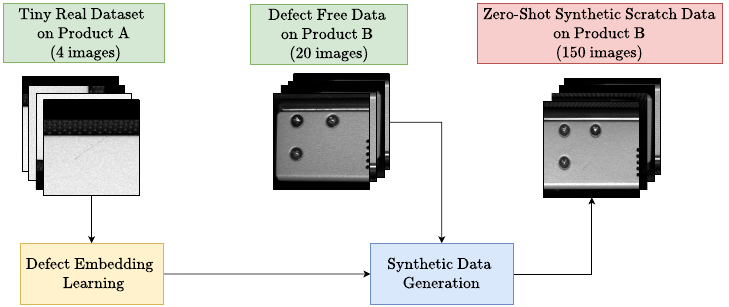}
    \caption{Flow for Zero-Shot Synthetic Data Generation: Capturing defect embeddings from 4 Product A samples to synthesize 150 high-fidelity defects directly onto Product B surfaces.}
    \label{fig:experiment_pipeline}
\end{figure}

The private industrial dataset is utilized once again. The target domain, Product B, consists of a gray metal surface with scratches, and the source domain, Product A, consists of a white metal surface with scratches. While a total of 133 real scratch samples on Product B surfaces exist, they are reserved exclusively for the evaluation of the models. Thus, it is strictly assumed that no real scratch data is available for Product B during training. Instead, 20 real defect-free images of Product B are utilized as backgrounds. From the Product A dataset, four real scratches are sampled (again selected as a representative value within the 3--5 image operational range) and their defect embeddings are learned using the proposed generative pipeline. Finally, 150 unique defect images are synthesized using the 20 backgrounds. The zero-shot synthetic data generation process is depicted in Figure~\ref{fig:experiment_pipeline}. To evaluate this scenario, three distinct model configurations are investigated as shown in Table~\ref{tab:zero_shot_configurations}. The first is the Product A model, serving as the baseline, which is trained entirely on the 150 real scratch samples from Product A. The second is the Product B synthetic model, which is trained exclusively on the 150 generated images for the Product B surface to evaluate pure zero-shot capabilities. The third is the Product B Synthetic Finetuning Model, which takes the trained baseline model, Product A Model, and fine-tunes it using the synthetic Product B data to assess the benefits of synthetic domain adaptation. 

\begin{figure}[h]
    \centering
    \includegraphics[width=1\textwidth]{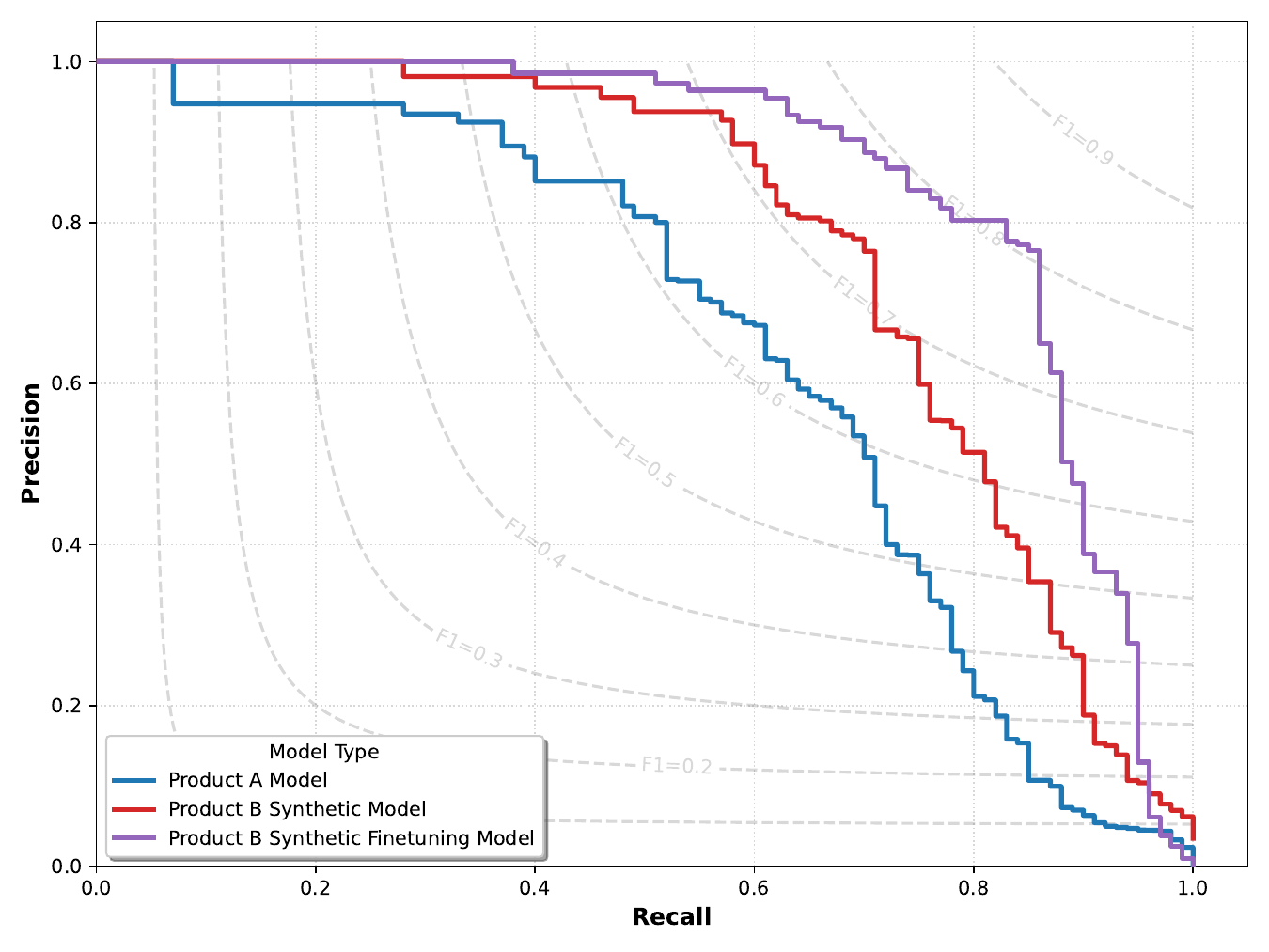}
    \caption{Precision-Recall curves for Zero-Shot detection experiment models.}
    \label{fig:pr_curve_zero}
\end{figure}

\begin{table}[h]
    \centering
    \caption{Zero-Shot Detection Experiment Performances at IoU@0.01 and Operating Point (Confidence Threshold of 40\%)}
    \label{tab:zero_shot_metrics}
    \renewcommand{\arraystretch}{1.2}
    \begin{tabular}{l c c c c c}
        \hline
        & \multicolumn{2}{c}{Overall Performance} & \multicolumn{3}{c}{Operating Point (40\%)} \\
        \cmidrule(lr){2-3} \cmidrule(lr){4-6}
        Model Name & mAP@0.01 & Best F1 & Precision & Recall & F-1 \\
        \hline
        Product A Model & 65.0\% & 63.5\% & 67.5\% & 59.4\% & 63.2\% \\
        Product B Synthetic Model & 77.7\% & 73.4\% & \textbf{93.8\%} & 54.9\% & 69.3\% \\
        Product B Synthetic Finetuning Model & \textbf{85.1\%} & \textbf{81.5\%} & 92.6\% & \textbf{65.4\%} & \textbf{76.7\%} \\
        \hline
    \end{tabular}
\end{table}

As shown in Figure~\ref{fig:pr_curve_zero} and Table~\ref{tab:zero_shot_metrics}, the Product A Model struggles to generalize to the new Product B surface, as it has never encountered the specific textural features of the gray metal background. This lack of domain awareness makes it difficult for the detector to differentiate between subtle surface patterns and real scratches, leading to a low baseline mAP of 65.0\% and a best F1-Score of 63.5\%.

Conversely, the Product B Synthetic Model achieves 77.7\% mAP and a best F1-Score of 73.4\%. This performance gap indicates that the synthetic data generated by the proposed pipeline provides domain-specific information that real data from the Product A source domain cannot supply. The experiments also highlight a critical stability requirement: positive defect samples are essential when adapting a model to a new background. This hypothesis was tested by retraining the baseline model on exclusively defect-free background images, which triggered catastrophic forgetting; the model lost its internal representation of defect structures while over-adapting to the new surface texture, resulting in a severe performance collapse.

The proposed generative pipeline addresses this stability gap. The Product B Synthetic Finetuning Model successfully adapts source knowledge using synthetic samples, yielding the highest performance across all metrics: 85.1\% mAP@0.01 and 81.5\% best F1-score. To assess operational utility, specific operating points on the Precision-Recall curves (Figure~\ref{fig:pr_curve_zero}) are highlighted, corresponding to a confidence threshold of 40\%. As detailed in Table~\ref{tab:zero_shot_metrics}, the Product B Synthetic Finetuning Model further pushes the operational performance to a 76.7\% F1-Score, supporting the transfer of defect detection capabilities to novel industrial surfaces.

While detection performance across models eventually tends to converge as real industrial data accumulates over extended production cycles, the primary impact of the proposed generative framework is the acceleration of the NPI timeline. By establishing a functional detection baseline before any real defect samples have been observed, the pipeline reduces the dependence on accumulated sample collection. This shifts manufacturing quality control toward a proactive paradigm where automated inspection systems are operationally effective from the very onset of production.

\section{Conclusion}

In this work, an end-to-end generative framework is presented for few-shot industrial defect synthesis designed to address one of the central bottlenecks in automated quality control: the scarcity of representative defect data during early deployment. By disentangling defect morphology from source-background appearance, the proposed pipeline enables high-fidelity synthesis of localized defects directly on target surfaces and supports both few-shot augmentation and zero-shot domain adaptation.

Experimental results indicate that the framework delivers visual and quantitative benefits in both operational scenarios. In the few-shot setting, augmenting a minimal real dataset with synthetic samples increased detection performance from 78.8\% to 83.3\% mAP@0.01. In the zero-shot setting, adapting source-domain knowledge through synthetic target-domain data improved performance from 65.0\% to 85.1\% mAP@0.01. These results suggest that realistic defect synthesis can serve not merely as a supplementary augmentation strategy, but as a practical mechanism for establishing strong inspection baselines before sufficient real defect data is available.

From an industrial perspective, the primary significance of the framework lies in its potential to shorten the NPI timeline. Rather than waiting for defects to accumulate naturally on a new line, manufacturers can begin with a small number of examples, generate domain-relevant training data, and deploy more effective inspection models at the outset of production. In this sense, the proposed approach supports a shift from reactive quality control toward a more proactive inspection paradigm.

At the same time, several limitations remain. The current evaluation is focused on scratch defects using an industrial partner's dataset, representing a standard setting for deploying such custom inspection systems. In addition, although the synthetic-only setting performs strongly, a residual gap remains between purely synthetic and mixed-data training, indicating that further improvements in diversity and appearance realism are still needed. Future work will focus on expanding to broader defect categories by learning and improving the quality of defects with higher structure, such as text defects, and increasing the harmony and adherence of the color and structure of the generated defects to the target surface. Furthermore, future efforts will explore creating a defect library and a system where, given a defect-free surface, possible defects are automatically retrieved from the library. This system will then synthesize these defects to create a realistic simulation dataset, enabling the training of a general defect detector using only defect-free surface data.

Overall, the results support the view that few-shot generative defect synthesis is a promising direction for industrial inspection, particularly in scenarios where deployment speed and data scarcity are tightly coupled. This framework offers a practical foundation for accelerating inspection readiness in real manufacturing environments.

\section*{Declaration on Generative AI}
During the preparation of this work, the authors used Large Language Models for language refinement and proofreading, and coding agents to support the implementation of the generative pipeline. After using these tools, the authors critically reviewed and edited the output. The authors take full responsibility for the scientific ideas, methodology, results, and conclusions presented in this manuscript and maintain full accountability for its integrity.

\section*{Acknowledgement}
The authors acknowledge the support of the Scientific and Technological Research Council of T\"urkiye (T\"UB\.ITAK) through the B\.IDEB 2224-A Grant Programme for the participation and conference presentation expenses of the first author, Serkan Hamdi G\"u\u{g}\"ul.

\end{document}